# Exposure and Emergence in Usage-Based Grammar:

# Computational Experiments in 35 Languages

*Draft Manuscript*

Jonathan Dunn
jonathan.dun@canterbury.ac.nz
(ORCID 0000-0002-1189-1908)

Department of Linguistics
and
New Zealand Institute for Language, Brain and Behaviour
University of Canterbury
Christchurch, New Zealand

**_Abstract_**

This paper uses computational experiments to explore the role of exposure in the emergence of construction grammars. While usage-based grammars are hypothesized to depend on a learner's exposure to actual language use, the mechanisms of such exposure have only been studied in a few constructions in isolation. This paper experiments with (i) the growth rate of the constructicon, (ii) the convergence rate of grammars exposed to independent registers, and (iii) the rate at which constructions are forgotten when they have not been recently observed. These experiments show that the lexicon grows more quickly than the grammar and that the growth rate of the grammar is not dependent on the growth rate of the lexicon. At the same time, register-specific grammars converge onto more similar constructions as the amount of exposure increases. This means that the influence of specific registers becomes less important as exposure increases. Finally, the rate at which constructions are forgotten when they have not been recently observed mirrors the growth rate of the constructicon. This paper thus presents a computational model of usage-based grammar that includes both the emergence and the unentrenchment of constructions.

# 1 Exposure and Emergence in Usage-Based Grammar

A central idea in usage-based grammar is that language forms a complex adaptive system that emerges given exposure to usage (Beckner et al. 2009; Bybee 2006; Divjak 2019). This means that a language learner is expected to be strongly influenced by observed frequencies and co-occurrence probabilities: of words, of phrases, and, ultimately, of constructions. But how do we evaluate this usage-based hypothesis without arbitrarily selecting a few constructions in isolation to experiment with? The most straight-forward approach is to predict grammars given exposure and then evaluate the quality of those grammars.

This paper simulates the emergence of usage-based grammars by learning constructicons for 35 languages (CxGs: Goldberg 2006). A constructional approach to language focuses on symbolic form-meaning mappings that are potentially idiomatic. Usage-based CxG further focuses on how constructions emerge given exposure and how they compete with other potential constructions. This paper simulates emergence and competition given exposure using computational CxG (Dunn 2017), a specific theory of usage-based construction grammar that predicts a constructicon given the exposure contained in a corpus. Computational CxG simulates the emergence of grammars by producing grammars. For the purposes of this paper, *CxG* refers to the general paradigm of Construction Grammar (Goldberg 2006). We use the term *constructicon* to refer to a specific grammar (an instance of a CxG) and the term *construction* to refer to a specific structure within that grammar.

The experiments in this paper apply computational CxG to comparable corpora for 35 languages. These corpora are drawn from three distinct registers in order to capture the influence of register-specific exposure. The _first_ experiment measures the growth of the constructicon as it is exposed to more usage. How does the growth of the constructicon compare to the growth of the lexicon across languages and across registers? The _second_ experiment measures the convergence of grammars that are exposed to different sets of usage from register-specific corpora. How much exposure is required before register-specific grammars converge onto the same representations? The _third_ experiment measures the unentrenching of constructions when the learner has not been recently exposed to them. This experiment uses a model of incremental exposure from large corpora to explore how grammars are pruned over time as constructions are forgotten. Taken together, these three experiments provide a robust model of the role of exposure in the emergence of usage-based construction grammar.

These experiments simulate the acquisition of constructions by incrementally increasing the amount of exposure: 100k words, 200k words, 300k words and so on up to 2 million words (c.f., Alishahi and Stevenson 2008; Beekhuizen et al. 2015; Matusevych et al. 2013). This provides a series of grammars that each represent an increasing amount of exposure. This basic framework is repeated across 35 languages and three registers. Within each language, these separate registers represent different sub-sets of exposure that are associated with a specific context of production. These parallel corpora thus allow us to experiment across both the amount and the type of exposure within each language.

The simulations or experiments in this paper are purely corpus-based in that there is no access to embodied experience or to non-distributional representations of meaning. Thus, these simulations are not a psycholinguistic observation of the emergence of CxGs in a specific set of individual participants. With access to individual participants, we could observe both production and perception together (with perception being dependent on the production of others). Instead, these simulations are a perception-based model of exposure. As described further in Section 4, the registers observed in this paper contain fluent adult usage, thus not approximating the exact exposure that would take place during L1 acquisition. The advantage of abstracting away from specific language learners in this way is scope: we are able to simulate the emergence of structure across an entire grammar in a highly multi-lingual and cross-register setting. The disadvantage is that the results of these experiments do not reflect the learning process of specific individuals and must be synthesized with participant-based studies in order to provide a more complete picture.

We start by reviewing previous work in computational CxG and how it relates to non-computational work (Section 2). The implemented theory of usage-based CxG is then described in detail (Section 3), followed by the data and methods used for carrying out the experiments in this paper (Section 4). The main experiments are then presented: growth rates or growth curves (Section 5), convergence rates (Section 6), and unentrenchment rates (Section 7).

The experiments in this paper provide a full-scale theory of usage-based construction grammar that predicts constructicons given exposure to actual usage. The growth of the constructicon is shown to be significantly slower than the growth of the lexicon (Section 5), a fact which results from the constructicon acquiring more general representations as the amount of exposure increases (Section 6). Finally, the rate at which constructions become unentrenched mirrors the rate at which constructions emerge (Section 7). This means that the role of exposure in grammar is an on-going process that continues even after the core constructicon has been learned.

## 2 Comparing Computational and Non-Computational CxG

Previous work on computational CxG has explored how to discover potential constructions (Dunn 2017; Forsberg et al. 2014; Wible and Tsao 2010, 2017). This same problem of finding constructions has also been approached using collostructional analysis (Hampe 2011) and using pattern-based sequences (Brezina et al. 2015; Hunston 2019; Martí et al. 2018; Perek and Patten 2019). Other computational research has simulated the process of learning constructions within specific pattern types (Barak and Goldberg 2017; Barak et al. 2017) and probed whether constructions are implicitly encoded in language models such as BERT (Tayyar Madabushi et al. 2020).

Most work on computational CxG has taken a usage-based approach, for example explicitly modelling the trade-off between memory and computation (Dunn 2018a). Usage-based models have also supported computational experiments on whether exposure is best described using frequency or association (Dunn 2019a; c.f., non-computational experiments in Flach 2020). A few other computational approaches, like Fluid Construction Grammar (Van Trijp 2015), are knowledge-based rather than usage-based. These approaches require a linguist to manually define each construction. In addition to being difficult to scale and unreproducible, these grammars are not able to answer experimental questions about exposure and the competition between constructions.

A recent line of computational work has leveraged usage-based construction grammar to describe syntactic variation across dialects (Dunn 2018b, 2019b, 2019c; Dunn and Wong 2022). This work models the competition between constructions within a wider geographic community, the idea being that some variants are preferred by a given community and are thus able to predict membership in that community. The effectiveness of constructicons for predicting dialect membership provides an external evaluation of grammar quality: better constructicons should provide better descriptions of dialectal variation. Other recent work has focused on individual differences (Dunn and Nini 2021; c.f., Anthonissen 2020) and on register differences (Dunn and Tayyar Madabushi 2021). This body of research in computational CxG shows that it is able to model variation in usage, whether that variation is caused by dialects or registers or individual differences. From a theoretical perspective, both dialectal and individual variation are related to exposure: those producing different grammars have also learned given exposure to different grammars.

Outside of computational CxG, the influence of exposure on the emergence of constructions has been studied using methods that focus on a small part of the grammar, usually one or two constructions in isolation. For example, this kind of work provides an in-depth examination of a small number of constructions using either corpus-assisted or participant-based methods (c.f., Azazil 2020; Desagulier 2015; Sommerer and Baumann 2021; Theakston et al. 2012; Ungerer 2021). These in-depth studies are insightful in combining corpus-based and participant-based evidence. The gap that remains in our understanding of the constructicon is the very large network of actual and potential constructions to which speakers are exposed.

One of the core challenges for computational CxG is modelling competition between *potential* constructions: any given utterance could have been produced by a number of a slightly different constructional representations. The participant-based view of this problem is a matter of production: why and when do speakers produce one potential construction over another? Recent work, for example, has found that models with access to information about specific lexical items perform worse at predicting which alternate construction will be produced (Liu and Ambridge 2021). A different line of work has focused on the means by which some (potential) constructions pre-empt one another in production (Goldberg 2011, 2016).

The difference between (i) models of competition between constructions in production and (ii) models of the emergence of grammars from corpora is a matter of scale: a few constructions in isolation have many fewer pre-emptive relationships than the network of all possible constructions. The challenge in this paper, and for the usage-based hypothesis more broadly, is to make predictions given the full hypothesis space of potential grammars. As mentioned before, the scale provided by computational models comes at the cost of limiting our scope to exposure that is contained in a corpus. This makes it difficult to incorporate the influence of stimuli from outside a corpus (i.e., Perek and Goldberg 2017).

The experiments in this paper use corpora from distinct registers to represent distinct sets of exposure. A substantial body of research has shown that register is a major source of linguistic variation in corpora (Biber 2012). Recent work has shown that register causes more variation in corpora (and thus in exposure) than even geographic dialects (Dunn 2021). These register-specific corpora are taken from social media, the web, and Wikipedia (Dunn 2020; Dunn and Adams 2020), providing comparable corpora for each of the 35 languages (c.f., Section 4).

## 3 Computational Construction Grammar

Computational CxG is a theory in the form of a grammar induction algorithm that provides a reproducible constructicon given a corpus of exposure (Dunn 2017). The theory is divided into three components, each of which provides a model of a particular aspect of the emergence of constructicons given exposure. This section briefly reviews each of these components.

*First*, a psychologically-plausible measure of association, the ΔP, is used to measure the entrenchment of potential constructions (Dunn 2018c; Ellis 2007; Gries 2013). These potential constructions are sequences of lexical, syntactic, and semantic slot-constraints.

*Second*, an association-based beam search is used to identify constructions of arbitrary length by finding the most entrenched representations in reference to a matrix of ΔP values (Dunn 2019a). A beam search here is a parsing strategy that avoids relying on heuristic frames and templates for producing potential constructions.

*Third*, a Minimum Description Length measure is used as an optimization function that balances the trade-off between increased storage of item-specific constructions and increased computation of generalized constructions (Dunn 2018a). The idea here is that any construction could become entrenched but more idiomatic constructions come at a higher cost. An optimization function or loss function is a measure of model quality that is used to guide the learning process.

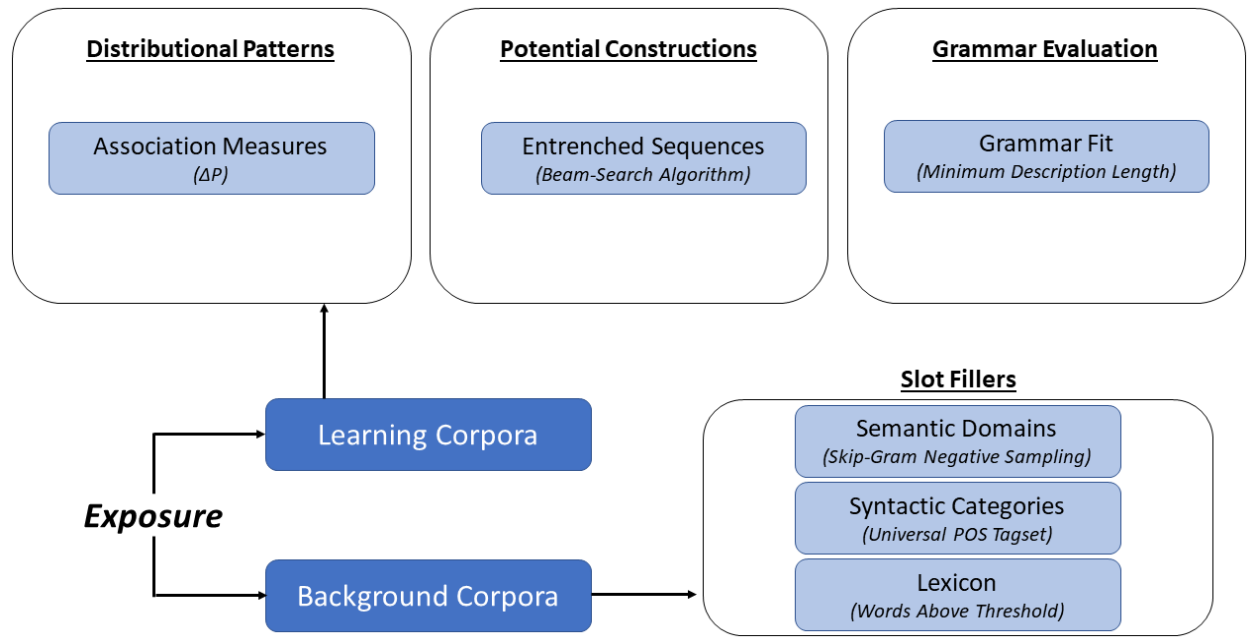

*Figure 1. Overview of the CxG Model Pipeline*

Given the complexity of the computational model, we show an overview of the approach in Figure 1. As shown in the bottom left of the figure, *exposure* is represented using corpus data for each language. This exposure is divided into two types: background corpora (used for forming distribution-based categories for the purpose of learning an inventory of possible slot-fillers) and learning corpora (used for the emergence of specific grammars). Thus, there is a distinction made, partly practical but partly theoretical, between the formation of basic categorizations within a language and the lexicalization of those categories within specific constructions. The computational model here is also a theory of CxG. As shown in the top of the figure, this model captures three separate stages in the emergence of grammatical structure: first, distributional patterns of words and sequences (these are based on strings and thus can be observed without any hypothesized syntactic structure); second, the identification of potential constructions as hypothesized structures that a learner might evaluate to explain the observed data; third, the evaluation of grammars that are selected from these potential constructions (where grammar quality is quantified using the fit between the grammar and the observed data). For the sake of clarity, Appendix 1 contains a glossary of terms used in this section.

Constructions here are constraint-based representations in which individual slots are limited to particular syntactic, semantic, or lexical fillers. CxGs are *usage-based* in the sense that constructions range from very idiomatic phrases (like the lexical constraints in 1a) to very abstract sequences (like the syntactic constraints in 1b). In computational CxG, the range of idiomaticity or productivity is implicitly captured by slot-constraints with varying abstractness. Constructions are represented as a sequence of slot-constraints, where each slot is defined by both the type of constraint and the specific value of the constraint. For example, the construction in (1a) would be a child of the construction in (1b). This coverage of both idiomatic and generalized constructions allows the grammar to capture variation across dialects, registers, and individuals as discussed above. But the challenge of allowing constructions with a range of abstractness is that the model must learn both the inventory of slot-fillers (i.e., semantic categories) as well as the slot-constraints themselves.

>    (1a) [ LEX: *give* – LEX: *me* – LEX: *a* – LEX: *hand* ]
>
>    (1b) [ SYN: VERB – SYN: NP – SYN: NP]
>
>    (1b') "sell us books"
>
>    (1b'') "trade them apples"

The first problem for computational CxG is to define the types of fillers that are used for slot-constraints. In this implementation, lexical constraints are based on word-forms, without lemmatization. These are the simplest and most idiomatic types of constraints. Syntactic constraints are formulated using the universal part-of-speech tagset (Petrov et al. 2012) and implemented using the Ripple Down Rules algorithm (Nguyen 2016). Semantic constraints are based on distributional semantics, with k-means clustering applied to discretize pre-trained fastText embeddings (Grave et al. 2019). These are character-based embeddings created by the skip-gram with negative sampling algorithm (Mikolov et al. 2013). Because of instability in embeddings given relatively small corpora and the impact of register on the resulting semantic domains (Dunn et al. 2022), we train a single set of domains that is shared across all grammars regardless of the specific exposure condition. The semantic constraints in the examples below are formulated using the index of the corresponding clusters, where each cluster is a discrete group of words from a single semantic domain. A complete list of semantic domains used in this paper, along with a grammar for each language, is available in the supplementary material.[1] Slot-fillers are not required to be constituents because constructions often form catena sequences instead (Osborne and Gross 2012).

Each sentence in the input corpus is represented using these three parallel types of slot-constraints (lexical, syntactic, semantic). In the first stage of the model, a co-occurrence matrix is produced that represents the association between all pairs of fillers using the ΔP measure, shown below. The measure, calculated for the left-to-right variant, is the probability that two units occur together (*X* and *Y*) adjusted by the probability that *X* occurs alone. In this notation, $Y_P$ indicates that unit *Y* is present and $Y_A$ that unit *Y* is absent. The subscript LR indicates that this describes only the left-to-right variant (the right-to-left variant is described further in Dunn 2018c). What distinguishes the ΔP from more common measures like Positive Pointwise Mutual Information (PPMI: Church and Hanks 1989; Dagan et al. 1993) is that it has direction-specific variants that take ordering into account, thus helping to capture syntactic patterns.

$$\Delta P_{LR} = p(X_P|Y_P) - p(X_P|Y_A)$$

Given (i) the three types of slot-constraints for each sentence and (ii) a co-occurrence matrix with the directional association for each pair of slot-fillers, a beam search algorithm is used to find the most *entrenched* sequences of constraints in the training corpus. The basic idea behind this search is to traverse all possible paths of slot-constraints, ending each path when the cumulative ΔP falls below a threshold (Dunn 2019a). For each unique span, the sequence of slot-constraints with the highest cumulative association is added to the provisional grammar. These constraint sequences constitute *potential* constructions. The beam search thus avoids fixed length templates (c.f., Dunn 2017) by incorporating parsing methods into the search for associated sequences. By exploring all possible paths in a systematic manner, the algorithm provides more robust representations while not producing a large number of noisy sequences that must be manually discarded.

---

[1] https://doi.org/10.18710/CES0L8

Pseudo-code for the association-based algorithm is shown in Table 1; this algorithm views a construction as a sequence of slot-constraints and uses the total directional ΔP to evaluate potential constructions. The search follows transitions from one slot-constraint to the next, proceeding left-to-right through the sentence; a *transition* here is the relationship between adjacent slot-constraints. Any transition below a threshold ΔP stops that line of the search. This algorithm references both local and construction-wide association values.

Any potential construction identified by this search whose transitions exceed the ΔP threshold is added to the candidate stack (the hypothesis space of potential constructions). At the end of the search, this stack is scored using each candidate's total ΔP across all slot-constraints. While primarily a transition-based parsing algorithm, this approach thus incorporates some global evaluation methods (c.f., Nivre and McDonald 2008; Zhang and Nivre 2012).

Table 1. Beam Search for Finding Potential Constructions (c.f., Dunn, 2019a)

| **Variables** |
| --- |
| *node* = unit (i.e., word) in line |
| *startingNode* = start of potential construction |
| *state* = type of slot-constraint for node |
| *path* = route from root to successor states |
| [*c*] = list of immediate successor states |
| $c_i$, $c_i+1$ = transition to successor constraint |
| *candidateStack* = hypothesis space of plausible constructions |
| *evaluate* = maximize $\sum$ΔP for $c_i$, $c_i+1$ in *path* |
| **Main Loop** |
| for each possible startingNode in line: |
|     RecursiveSearch(path = startingNode) |
| evaluate candidateStack |
| **Recursive Function** |
| RecursiveSearch(path): |
|     for $c_i$, $c_i+1$ in [c] from path: |
|         if ΔP of $c_i$, $c_i+1$ > threshold: |
|             add $c_i+1$ to path |
|             RecursiveSearch(path) |
|         else if path is long enough: |
|             add to candidateStack |

Given a hypothesis space of potential constructions, which the learner has been exposed to, the final component models the entrenchment of these potential constructions. This model is based on the Minimum Description Length paradigm (Goldsmith, 2001, 2006; Grünwald, 2007). In this kind of model, observed probabilities are used to calculate the encoding size of a grammar (the $L_1$ term) as well as the encoding size of a test corpus given that grammar (the $L_2$ term). Usage-based grammar posits a trade-off between memory and computation; this is modelled by MDL's combination of $L_1$ and $L_2$ encoding size. The first is a measure of the complexity of the constructicon itself and the second is a measure of the fit between the constructicon and a given test corpus.

$$MDL = \min_{G}\{L_1(G) + L_2(D \mid G)\}$$

The best grammar is the one which minimizes this metric on a test corpus. In the equation above, *G* refers to the grammar being evaluated and *D* refers to the test corpus (or data set). For example, this is used to choose the parameters of the beam search algorithm described above. In practice, the use of MDL to evaluate grammars is quite similar to the use of perplexity to evaluate language models: both are specific to each test corpus because they describe the fit between the model and the data (and c.f. Goldsmith 2015, for viewing this from a linguistic perspective). The advantage of the MDL metric for usage-based grammar is that it distinguishes between the complexity of the grammar ($L_1$) and the fit between the grammar and the test corpus ($L_2$). In short, this approach implements the idea that any potential construction can become entrenched but that not all potential constructions are worth storing.

This grammar induction algorithm constitutes a computational theory of CxG which predicts a constructicon given a corpus that represents exposure. This constructicon is a set of constructions, each of which is a sequence of slot-constraints. And each slot-constraint is formulated using the basic inventory of lexical, syntactic, and semantic fillers. The examples below show different types of constructions; the full constructicons for each language with examples are provided in the supplementary material.[2] In these representations, each slot-constraint is separated by a dash ("–") and the construction is enclosed in brackets. The types of slot-constraint are abbreviated: Lexical (LEX), Syntactic (SYN), and Semantic (SEM). The fillers for each slot-constraints are produced as described above. Below each construction are five examples or tokens of that construction, taken from the Wikipedia corpus.

The first several examples show noun phrases and adposition phrases of varying levels of abstraction. In (2) we see an idiomatic adposition phrase with partial lexical constraints. This construction is idiomatic both because it contains lexical constraints but also because it is relatively fixed: there is limited variation in its tokens. We can contrast this with the more abstract noun phrase in (3), which contains all three types of slot-constraints. The final constraint is semantically defined (with the number referring to the distributional cluster). The construction thus combines a specific form and a specific semantic domain.

> (2) [ LEX: *due* – SYN: ADP – SYN: DET – SYN: NOUN ]
> (2a) "due to this difference"
> (2b) "due to the time"
> (2c) "due to the fact"
> (2d) "due to the risk"
> (2e) "due to the origin"
>
> (3) [ SYN: NOUN – LEX: *of* – SYN: DET – SEM: <587> ]
> (3a) "part of the frontier"
> (3b) "possibility of a utopia"

---
[2] https://doi.org/10.18710/CES0L8

(3c) "provinces of the empire"
(3d) "part of the dominion"
(3e) "duty of the crown"

In (4), we again see a semantic constraint, this time on the adjective within a noun phrase. Here we see that a semantic constraint on a modifier also serves as an implicit semantic constraint on the head. A longer construction is given in (5), now containing only syntactic and semantic constraints, with the semantic constraint reserved for the final noun. Again, a semantic constraint in one slot has an implicit influence on all other slots.

(4) [ SYN: ADJ – SEM: <590> – SYN: NOUN ]
(4a) "exclusive anarchist schools"
(4b) "radical left-wing ideology"
(4c) "revolutionary socialist direction"
(4d) "French socialist movement"
(4e) "Mexican liberal party"

(5) [ SYN: NOUN – SYN: ADP – SYN: DET – SYN: ADJ – SYN: <536> ]
(5a) "impact on the bigger currents"
(5b) "coastline on the Atlantic ocean"
(5c) "systems of the Pacific ocean"
(5d) "outlet in the Arctic ocean"
(5e) "shore of the African landmass"

Sample verb phrase constructions are given in (6) and (7). In the first case, the subject is also included and there is a lexical constraint (*is*) rather than a syntactic constraint (AUX). This potential variation shows the impact of competing slot-constraints: there are many slightly different but closely related formulations of constraints. Because usage-based grammar is specific to a set of exposure (here the training corpus), we as linguists cannot know given our own intuitions which of these candidates is the entrenched form. A transitive verb phrase is represented in (7), here with a lexical constraint for the head verb. As before, an idiomatic constraint in one slot has an implicit influence on the other slots: here, the argument of "write" is limited based on the lexical constraint for that verb.

(6) [ SYN: NOUN – LEX: *is* – SYN: ADV – SYN: VERB ]
(6a) "anarchism is often considered"
(6b) "advertising is somewhat restricted"
(6c) "anthropology is typically divided"
(6d) "alchemy is sometimes associated"
(6e) "transmutation is routinely performed"

(7) [ LEX: *write* – SYN: DET – SYN: NOUN ]
(7a) "write the screenplay"
(7b) "write a book"
(7c) "write the tune"
(7d) "write an argument"
(7e) "write a paper"

We also see constructions larger than a single verb phrase in (8) and (9), both defined using mostly syntactic constraints. In the first case, there is a verb with an adpositional argument and a coordinating conjunction. The presence of this conjunction means that the construction does not capture an entire constituent, instead forming a catena. In (9), we see a sentence level construction that could be informally represented as "X is the Y of Z".

> (8) [ SYN: VERB – SYN: ADP – SYN: DET – SYN: NOUN – SYN: CCONJ ]
> (8a) "leading to the conquest and"
> (8b) "forced upon the pupil and"
> (8c) "hides in the bushes and"
> (8d) "met at a party and"
> (8e) "defeated on the battlefield but"
>
> (9) [ SYN: NOUN – LEX: *is* – SYN: DET – SYN: NOUN – LEX: *of* – SYN: DET – SYN: NOUN ]
> (9a) "actuality is the fulfilment of the end"
> (9b) "telos is the principle of every change"
> (9c) "orchidaceae is the sister of the rest"
> (9d) "combinatorics is an example of a field"
> (9e) "Zan is the king of the accordion"

We see more idiomatic transitional phrases in (10) and (11), both containing all three types of slot-constraints. In (10), an adverbial clause is drawn from a specific semantic domain without capturing the object of the clause's verb. In (11), a formulaic phrase "in order to" is joined with a semantically-constrained verb.

> (10) [ LEX: *while* – SEM: <113> – SYN: ADP ]
> (10a) "while preparing for"
> (10b) "while working on"
> (10c) "while serving as"
> (10d) "while experimenting with"
> (10e) "while returning to"
>
> (11) [ LEX: *in* – LEX: *order* – SYN: PART – SEM: <583> ]
> (11a) "in order to assert"
> (11b) "in order to establish"
> (11c) "in order to provide"
> (11d) "in order to defend"
> (11e) "in order to formalize"

Finally, in (12) we see another sentence-level construction, this time a passive verb together with an agent contained in an adpositional phrase. But this is not simply the passive construction because the verb is constrained to a particular semantic domain and this, again, implicitly constrains the subject and the agent of that verb.

> (12) [ SYN: NOUN – LEX: *are* – SEM: <830> – LEX: *by* – SYN: DET – SYN: NOUN ]
> (12a) "GMOs are questioned by some ecologists"
> (12b) "nouns are negated by the particle"

(12c) "abbots are appointed by the brothers"
(12d) "specialists are hired by the client"
(12e) "waves are produced by the vibrations"

These examples show how the slot-constraints used to represent constructions allow for a range from formulaic phrases to fully productive sentences. Semantic and lexical constraints in one slot have an influence on all other slots in the construction. And we also see that there are many relatively minor permutations of slot-constraints that would have produced a rather similar set of tokens (like "the" vs. DET). The point here is to emphasize that the challenge for modelling the emergence of constructions is to capture the competition between potential constructions and the implicit relationships between slot-constraints within a construction. The hypothesis space, given a decent amount of exposure, is quite large.

How do we evaluate the quality of a predicted constructicon? It is clear that the intuition of a linguist is not a helpful standard, because usage-based grammar is a mapping from exposure to entrenchment. Thus, the exposure represented in a corpus is entirely independent of the intuitions of an individual linguist. Further, because constructions could contain catena (Osborne and Gross 2012), even the application of constituency tests to individual slot-constraints would not be helpful. Previous work has focused on both *internal* and *external* evaluations of computational grammars. An internal evaluation measures the fit of the grammar against a test corpus and then evaluates the relative quality of grammars which represent different hypotheses. An external evaluation uses a CxG for some type of analysis, for example modelling the differences between dialects. Better grammars will be better able to identify the dialects of different geographic communities.

The experiments in this paper can be seen as a different type of evaluation. This theory of CxG is not deterministic, so that it will produce somewhat different grammars potentially even on the same or similar data. Thus, experiments in growth rates and convergence rates can be used to find out how stable this approach is, both over the amount of exposure and over exposure drawn from distinct registers. A model that is not robust would see random fluctuations in both dimensions of comparison. Of course, the stability of a model is not the primary type of evaluation: very bad grammars could also be very stable. However, with previous work having conducted both internal and external evaluations, this paper focuses on these further tests of computational CxG.

## 4 Data and Methodology

The basic experimental approach in this paper is to learn grammars over increasing amounts of exposure: from 100k words to 2 million words in increments of 100k words (thus creating 20 grammars per language per register). This series of grammars simulates the accumulation of grammatical knowledge as the amount of exposure increases. This approach is repeated across each of the three registers that represent different contexts of production.

The register-specific data used to progressively learn grammars is collected from three sets of corpora: social media (TW for Twitter), non-fiction articles (WK for Wikipedia), and web pages (CC for Common Crawl), drawn from the *Corpus of Global Language Use* (Dunn 2020). This corpus contains the same amount of data per register per language, making each language-register pair comparable. The 35 languages used in the experiments are listed in Table 2, together with the three-letter code used to identify each language in the tables and figures to follow. The languages are sorted by family; most are from branches of Indo-European (23), but this leaves a significant number of non-Indo-European languages as well (12).

**Table 2. Languages by Family**

| Language | Code | Family | Language | Code | Family |
|---|---|---|---|---|---|
| Indonesian | ind | Austronesian | Catalan | cat | IE:Romance |
| Tagalog | tgl | Austronesian | French | fra | IE:Romance |
| Bulgarian | bul | IE:Balto-Slavic | Galician | glg | IE:Romance |
| Czech | ces | IE:Balto-Slavic | Italian | ita | IE:Romance |
| Latvian | lav | IE:Balto-Slavic | Portuguese | por | IE:Romance |
| Polish | pol | IE:Balto-Slavic | Romanian | ron | IE:Romance |
| Russian | rus | IE:Balto-Slavic | Spanish | spa | IE:Romance |
| Slovenian | slv | IE:Balto-Slavic | Arabic | ara | Semitic |
| Ukrainian | ukr | IE:Balto-Slavic | Hebrew | heb | Semitic |
| Danish | dan | IE:Germanic | Estonian | est | Uralic |
| German | deu | IE:Germanic | Finnish | fin | Uralic |
| English | eng | IE:Germanic | Hungarian | hun | Uralic |
| Dutch | nld | IE:Germanic | Greek | ell | Misc. |
| Norwegian | nor | IE:Germanic | Korean | kor | Misc. |
| Swedish | swe | IE:Germanic | Thai | tha | Misc. |
| Farsi | fas | IE:Indo-Iranian | Turkish | tur | Misc. |
| Hindi | hin | IE:Indo-Iranian | Vietnamese | vie | Misc. |
| Urdu | urd | IE:Indo-Iranian | | | |

The experiments here depend on limiting the amount of training data as a means of controlling for different levels of exposure. In each condition, then, the same training data is used for each stage in the algorithm. In other words, the 200k word exposure condition has access only to the 200k word training corpus (with the implicit exception of the pre-trained embeddings). Each model is trained on the same underlying corpus, so that the 500k word condition is given the same data as the 400k word condition plus an additional 100k words of new data. The result is that these experiments also account for stability of the underlying model: large differences between the constructicons resulting from 1.7 million words of exposure and 1.8 million words would be the result of arbitrary fluctuations in the algorithm itself. Recent work on the cross-linguistic stability of relationships between registers supports this approach of observing the same set of registers in a multi-lingual setting (Li et al. 2022).

## 5 Growth Rates of the Lexicon and Constructicon

The constructicon is hypothesized to be located on a continuum that connects grammar and the lexicon. While the growth of the lexicon is well-documented (Baayen, 2001; Gelbukh and Sidorov 2001; Zipf 1935; Heaps 1978), the growth of the constructicon has been studied much less, in part because previous work has been unable to predict the constructicon given a corpus of exposure. Previous computational work has focused on the growth of the constructicon in individuals vs. random aggregations of individuals (Dunn and Nini 2021), finding that those grammars exposed to individuals grow significantly more quickly. Note that the term *growth rate* is often used in the computational literature while the term *growth curve* is more common in the participant-based literature. The two can be used interchangeably for our purposes.

Given Heap's Law, the lexicon is expected to have a growth rate such that the number of vocabulary items in the corpus (i.e., exposure) is proportional to the exponent of the size of the corpus, as shown below. Here, *V* is the size of the vocabulary (words in the lexicon and constructions in the constructicon). *W* represents the number of words in the corpus and its exponent *α* represents the crucial parameter which describes the growth rate of the lexicon; *k* is a fixed constant.

$$V = k * W^\alpha$$

The challenge in measuring the growth rate is to estimate the parameter α. The simplest method is to undertake a least-squares regression using the log of the size of the corpus and the log of the size of the vocabulary, shown below (c.f., Gelbukh and Sidorov 2001); the constant *k* becomes the intercept of this regression. Given that this problem involves a power-law distribution, the least-squares method is potentially problematic because fluctuations in the most infrequent items can lead to a poor fit at certain portions of the curve (Clauset 2009). Figures in the supplementary material plot the actual growth rate and the modelled growth rate to show that a least-squared regression adequately models the data in this case.[3]

$$log(V) = \alpha * \log(W) + k$$

A sample growth curve is shown in Figure 2 for English, with the lexicon on the left and the grammar on the right. The x-axis represents the amount of exposure (up to 2 million words) and the y-axis represents the number of items (up to almost 100k lexical items). Within each sub-plot, the lines represent the three registers (CC, TW, WK). First, we see that the overall size of the lexicon is significantly larger and the grammar growth rate is more nearly linear than the lexicon growth rate, which slows after a certain amount of exposure. We will investigate the growth rates more formally below. Second, we see that register has a much smaller impact than the distinction between grammar and the lexicon. This is as we would expect because register here is a different source of exposure, in effect reproducing the experiment in three contexts. The growth rates for all languages are available in the supplementary material.

**Figure 2. Growth Rate for English**

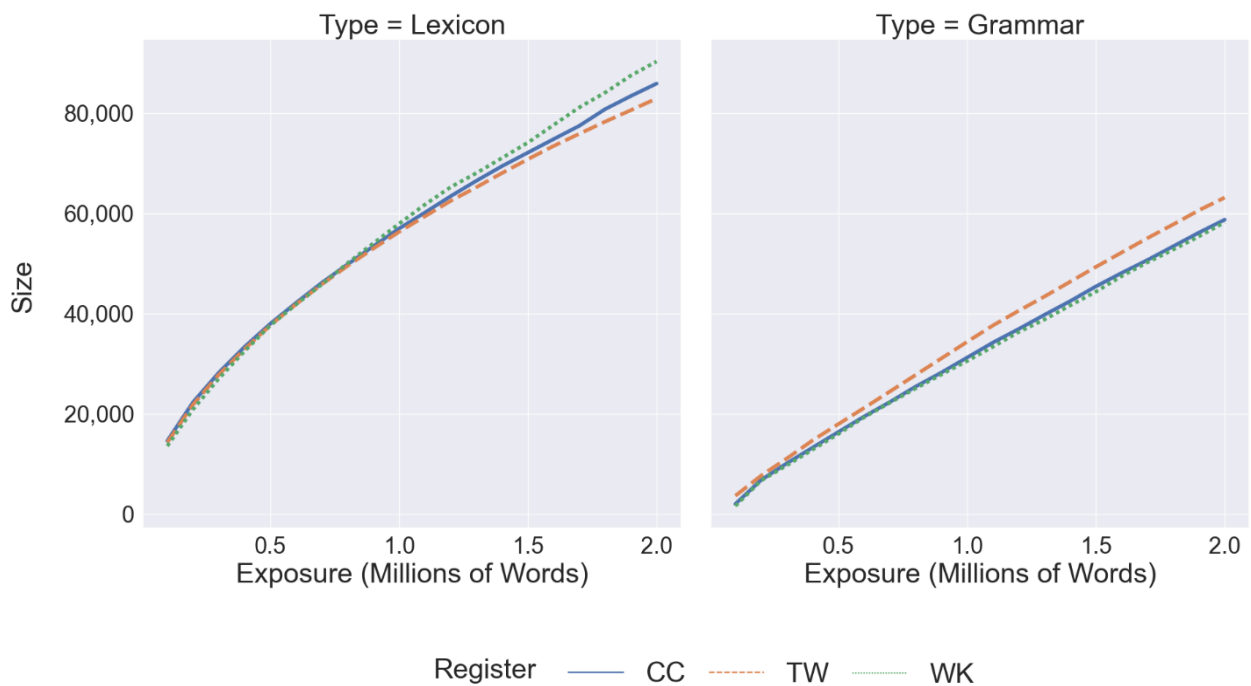

---

[3] https://doi.org/10.18710/CES0L8

We model growth rates using least-squares linear regression to estimate the α parameter (additional figures in the supplementary material show that this approach provides an adequate model of the observed growth). Given Heap's Law, the α represents the curve of the growth rate, with higher values indicating more linear growth rates. In other words, a high α indicates that the growth rate does not change given more exposure. To visualize the range of values, Figure 3 shows the lexicon curves for Korean (on the left) with a high α (mean = 0.772) and Urdu (on the right) with a low α (mean = 0.606). The scale is quite different for each language, because Korean ends up with many more lexical items overall. Thus, the Korean lexicon grows more quickly and its growth rate declines less over time.

**Figure 3. Growth Curve for High α (Korean) and Low α (Urdu)**

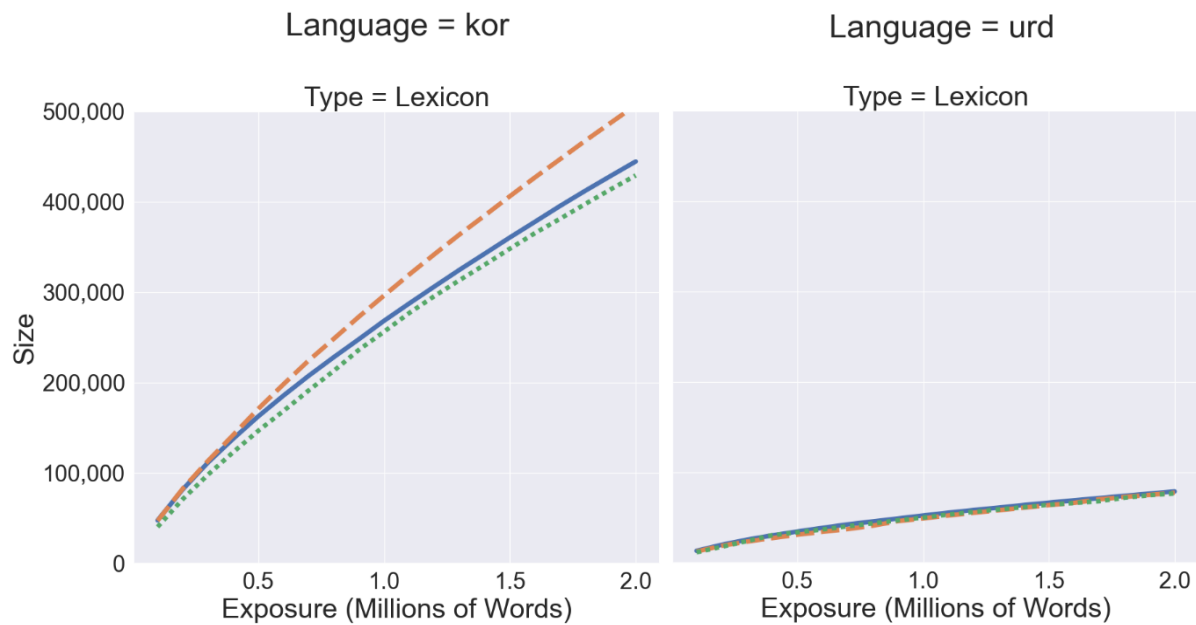

Figure 3 shows the difficulty of directly comparing the growth rates across languages: the lexicon does not include morphological parsing, for example, so that more agglutinative languages would have a higher growth rate simply as a side-effect of having more variations in word-forms. We control for this by comparing lexicon growth and grammar growth within languages. While the growth rate across very different languages is not necessarily comparable because of confounding factors like morphology, the growth rate between the grammar and the lexicon for the same language is always comparable.

The growth rates here correspond rather closely with Heap's Law (a corpus-based hypothesis), rather than with the sigmoid-like learning curve observed, for example, in age-of-acquisition studies (Kuperman et al. 2012). In part, this results from the means of observation: our assumption here is that a lexical item is learned at first exposure (a perception-based model) rather than at first observation in usage (a production-based model). In other words, do we count a lexical item as having been *learned* after perception or after production? While a learner may have passive knowledge at first exposure (our model here), the learner does not exhibit active knowledge until the lexical item has been observed in production (the Kuperman et al model). Thus, the difference in growth curves results from whether we model the learner's perception or whether we model the learner's production. In both cases the models have limitations: a computational model can completely control for exposure (perception), but is more limited in its ability to predict when passive understanding leads to active production. And a participant-based model can robustly observe the production of the learner, but is much more limited in its observation of the learner's complete exposure. A more detailed examination of this kind of scaffolded structure during learning remains a problem for future work.

Our first experimental question is whether the lexicon and constructicon actually grow at different rates. Here we use a 95% confidence interval for the least-squares regression used to estimate the parameter α. If the growth rates do not overlap given this confidence interval, there is a significant difference. It turns out that there is a significant difference in every case, with no exceptions. This means that in 105 out of 105 conditions (where each language-register pair is a condition) the lexicon and the grammar grow at significantly different rates. The full results for this experiment are available in the supplementary material.[4]

In all cases, the lexicon grows more quickly while the grammar has a higher α parameter (thus growing more steadily). A consistent difference is that the lexicon shows a larger burst of growth after the very first observation. In other words, the growth rate looks at the increase in size between 100k words of exposure and 200k words, and so on. But Table 3 shows the percent of the total inventory which is observed within the first 100k words (the values here are the mean across the three registers). For the lexicon, an average of 16.3% of the total items are observed within the very first exposure. But for the grammar this is only 4.5%. Thus, the grammar grows more evenly while the lexicon shows an initial burst of growth.

We might hypothesize that there is a lag in some sense between the growth of the lexicon and the growth of the grammar (c.f., Bates and Goodman 1997). These particular experiments would not allow us to measure such a lag because the lexicon used for learning a constructicon is derived from the same set of observations (rather than, for example, the previous set of observations in a time-lagged model). Future work will need to address this question, then, by simulating a time-lag in the lexicon available for determining potential constructions.

*Table 3. Initial Growth by Percentage of the Total Representations*

| *Language* | *Grammar* | *Lexicon* | *Language* | *Grammar* | *Lexicon* |
|---|---|---|---|---|---|
| ara | 4.8% | 16.8% | kor | 4.5% | 11.2% |
| bul | 4.4% | 18.1% | lav | 4.7% | 16.8% |
| cat | 3.1% | 18.6% | nld | 5.1% | 15.8% |
| ces | 4.7% | 16.9% | nor | 5.1% | 13.7% |
| dan | 6.0% | 12.9% | pol | 5.2% | 16.8% |
| deu | 5.1% | 14.3% | por | 5.0% | 20.7% |
| ell | 3.2% | 19.5% | ron | 5.1% | 18.6% |
| eng | 3.5% | 16.2% | rus | 5.0% | 16.0% |
| est | 5.1% | 13.0% | slv | 5.0% | 15.5% |
| fas | 5.2% | 15.6% | spa | 3.3% | 19.8% |
| fin | 4.5% | 13.0% | swe | 5.0% | 13.0% |
| fra | 5.5% | 19.6% | tgl | 4.7% | 14.9% |
| glg | 4.9% | 15.2% | tha | 4.6% | 25.0% |
| heb | 3.7% | 16.1% | tur | 4.9% | 16.2% |
| hin | 2.4% | 12.4% | ukr | 5.1% | 15.3% |
| hun | 5.0% | 14.2% | urd | 2.1% | 17.5% |
| ind | 4.9% | 17.6% | vie | 7.0% | 17.5% |
| ita | 5.4% | 19.8% | | | |

---

[4] https://doi.org/10.18710/CES0L8

Our second experiment looks at the relationship between the lexicon and the constructicon within each register: is the growth of the grammar entirely predictable given the growth in lexical items? In other words, if the growth in the constructicon is driven purely by an increase in the overall vocabulary, there should be a strong relationship across languages between the growth rates for each. We start by measuring the correlation between growth rates within each register. None of the three registers have a significant correlation between the growth of the lexicon and the growth of the grammar.

We visualize the relationship in Figure 4 using a regression plot within registers. The x-axis represents the lexicon's growth rate and the y-axis the constructicon's growth rate; each register is presented in a separate sub-plot. Each language provides a single observation. If the lexicon grows slowly, the language will fall toward the left; if the constructicon grows slowly, the language will fall toward the bottom. Thus, those languages in the bottom left have a relationship that shows low growth in both lexicon and constructicon. Few languages show either high or low growth in both: there is no trend across languages. This experiment shows that the grammar is not dependent on the lexicon to explain its growth given increased exposure. Together with the first experiment, this shows that the growth rates for the lexicon and the constructicon (i) do not overlap and (ii) do not have a causal relationship that could explain the growth of the constructicon.

**Figure 4. Relationship between Lexicon and Constructicon Growth Rates**

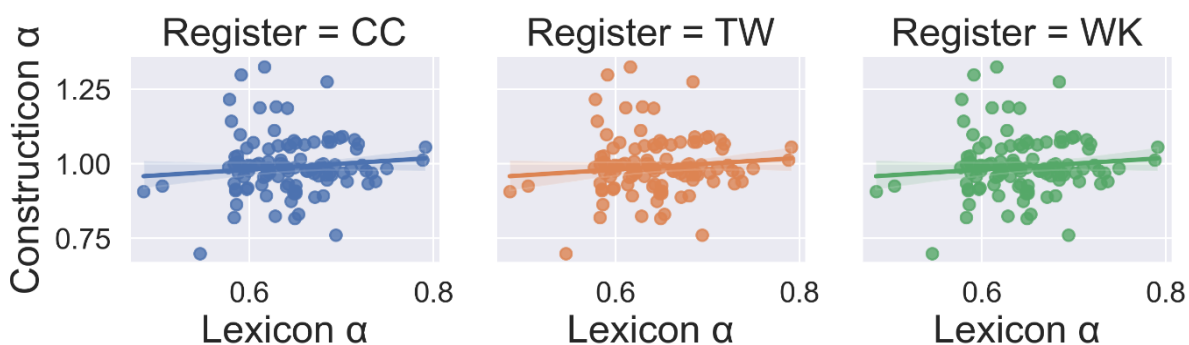

Our final question is whether related languages have similar growth rates: is language family a factor that can explain the varying growth rates of the constructicon? Language family encompasses other potential factors like morphological type or writing system. We estimate the mean and standard deviation for each family using a Bayesian confidence interval of 90%, as shown in Table 4. The lexicon is shown on the left and the constructicon on the right. As we know from previous experiments, the lexicon grows more quickly than the constructicon but also has a decline in growth rate given more exposure (i.e., it becomes sub-linear). We see variation across families in the growth of the lexicon: from 0.60 (Romance) to 0.71 (Uralic). But the range for the constructicon is much more constrained: from 0.97 (Germanic) to 1.04 (Indo-Iranian). The fact that the α parameter here is quite close to 1 means that that grammar grows at a linear rate, thus explaining why it behaves differently than the lexicon. This consistency across languages also means that language family is a factor in the growth of the lexicon but less so in the growth of the constructicon.

Table 4. Growth Rates by Mean and Std. Dev by Family

|  | Lexicon | | Constructicon | |
| ---: | :---: | :---: | :---: | :---: |
|  | *Mean* | *Std. Dev* | *Mean* | *Std. Dev* |
| **TOTAL** | **0.64** | **0.05** | **0.99** | **0.10** |
| Uralic | 0.71 | 0.02 | 0.99 | 0.05 |
| Balto-Slavic | 0.64 | 0.03 | 0.99 | 0.07 |
| Romance | 0.60 | 0.01 | 1.00 | 0.09 |
| Indo-Iranian | 0.64 | 0.04 | 1.04 | 0.22 |
| Austronesian | 0.62 | 0.05 | 1.02 | 0.16 |
| Semitic | 0.63 | 0.03 | 1.04 | 0.09 |
| Germanic | 0.66 | 0.04 | 0.97 | 0.06 |

The experiments in this section have shown that (i) there is always a significant difference in the growth of the lexicon and the constructicon, with the lexicon growing more quickly initially and overall and the grammar growing more steadily; and (ii) that there is not a cross-linguistic relationship between the growth rate of each. Further examination shows that language families share variation in the growth of the lexicon but not in the growth of the constructicon. The basic finding, then, is that the growth of the constructicon is not driven by the lexicon, is subject to much less variation across language families, and falls within a more constrained and linear range. The inclusion of three distinct registers means that the context of production is not a confounding factor that could have distorted these conclusions.

Does this finding call into question the idea in lexico-grammatical paradigms, like CxG, that there this no clear distinction between the lexicon and the grammar? On the one hand, these results do suggest that there is a distinction between simple constructions (with one slot) and complex constructions (with at least three slots). Especially in the case of syntactic and semantic constraints, there is both a matter of category formation and a matter of emerging structure. Thus, although there is a continuum between grammar and the lexicon, complex constructions depend on category formation much more than do simple (i.e., lexical) constructions.

Two further issues should also be noted here: first, each iteration of learning (at a certain amount of exposure) is independent in this implementation. However, given the observed lag between lexical acquisition and grammatical acquisition (Bates and Goodman 1997), a more precise model would define slot-constraints using categories from the previous model (i.e., a verb class learned at 500k words could only define slot-constraints in the next constructicon at 600k words). Second, the measure of grammar complexity used previously (c.f., Bates and Goodman 1997) is based on an annotated scale of sentence complexity while the measure here is based on the learned constructicon. As constructions become more complex they become more abstract. And more abstract constructions (like a phrase structure rule) cover more utterances than item-specific constructions (like a phrasal idiom); thus, fewer abstract constructions are needed. As a result, the growth curve within a constructicon should differ according to the level of abstraction. The growth curve of the grammar, then, involves both the number of representations but also the complexity of those representations. This relationship remains a question for further work.

## 6 Convergence Rates

Our next question is about the convergence rate for grammars of the same language that are exposed to different registers. For example, we learn a constructicon for Polish by exposing computational CxG to social media (TW), to non-fiction articles (WK), and to web pages (CC). For each level of exposure, we compare how similar the constructicons are. As exposure increases, do the grammars become more similar? Recent work has divided constructicons into core constructions (the most frequent) and peripheral constructions (the least frequent); this creates a proto-type structure in the grammar (Dunn and Tayyar Madabushi 2021). Working with Germanic and Romance languages, this work showed that full grammars steadily converge given increased exposure; but the core constructicon emerges early and does not change significantly. Here we take a closer look at this problem, with an experimental framework that includes more language families.

Our central measure of overlap between grammars is based on the Jaccard similarity, where values close to 1 represent very similar grammars and values close to 0 represent very different grammars. We adapt this similarity measure to support comparing constructicons as described below. This is a measure of set similarity: two constructicons are more similar when they contain more of the same constructions. This basic measure, shown below, is the ratio of the intersection of the sets over the union of the sets.

$$J(A, B) = \frac{|A \cap B|}{|A \cup B|}$$

The comparison between grammars is complicated by the large number of potential constructions. In other words, two different candidate constructions could use slightly different slot-constraints to capture a similar set of utterances, essentially providing two versions of the same construction. Consider the construction in (13), taken again from the English grammar and the corpus of non-fiction articles (WK). The tokens in (13a) through (13e) are examples of this construction. In each of these tokens, the particle is the infinitive particle "to" introducing an infinitive clause. This slot constraint could have been represented as a lexical constraint using "to" as shown in (13f) without changing any of the tokens of the construction. In practical terms, these two candidates are the same construction.

> (13) [ LEX: *has* – SYN: VERB – SYN: PART ]
> (13a) "has attempted to"
> (13b) "has continued to"
> (13c) "has failed to"
> (13d) "has begun to"
> (13e) "has learned to"
> (13f) [ LEX: *has* – SYN: VERB – LEX: *to* ]

Previous work has thus created a fuzzy Jaccard similarity, in which the definition of set membership is extended to very similar constructions. In this measure, a sub-sequence matching algorithm is used to find how many slot-constraints are shared between two constructions, taking order into account. Any two constructions above a threshold of 0.71 shared sub-sequences are considered a match. This threshold is chosen because it allows one slot-constraint to differ between most constructions while still considering them to be similar. For example, two six-slot constructions must share five constraints in order to count as a match at this threshold. This measure thus provides a better approximation of construction similarity, focusing on constructions with slightly different internal constraints or with an added slot-constraint in one position. This fuzzy measure is used to quantify the convergence between grammars across the entire constructicon.

A second measure, a frequency-weighted Jaccard similarity, focuses on the core constructions in the grammar. Here each construction is weighted using its frequency in an independent corpus. For each language, a background corpus is created from a mix of different registers: Open Subtitles and Global Voices (Tiedemann 2012) and Bible translations (Christodoulopoulos and Steedman 2015). This background corpus represents usage external to the three main registers used in the experiments. The frequency of each construction is derived from 500k words of this background corpus, so that very common constructions are given more weight in the similarity measure. The basic idea is that some constructions are part of the core grammar, thus being more frequent.

These two measures based on Jaccard similarity provide three values for each language: CC-WK, CC-TW, and WK-TW. Each value represents a pairwise similarity between two register-specific grammars. The higher these values, the more the learner is converging onto a shared grammar. The fuzzy measure captures convergence across the entire constructicon while the weighted measure focuses on the core constructicon. As before, we learn grammars from increasing amounts of data in order to capture change in convergence as the amount of exposure increases.

The convergence of grammars across different registers is shown in Figure 5 for English (left) and Spanish (right). The y-axis here is the Fuzzy Jaccard similarity, capturing shared constructions across the entire grammar. Values toward 1 indicate high overlap. The x-axis is the amount of exposure, from 100k words to 2 million words. Thus, as the figure moves to the right the model has been exposed to increasing amounts of data. Each of the three lines represents a different pair of registers: CC-TW, CC-WK, and WK-TW. Because we are examining the convergence of grammars when they are exposed to unique registers, higher values mean that the two constructicons are more similar.

**Figure 5. Convergence of the Constructicon for English and Spanish**

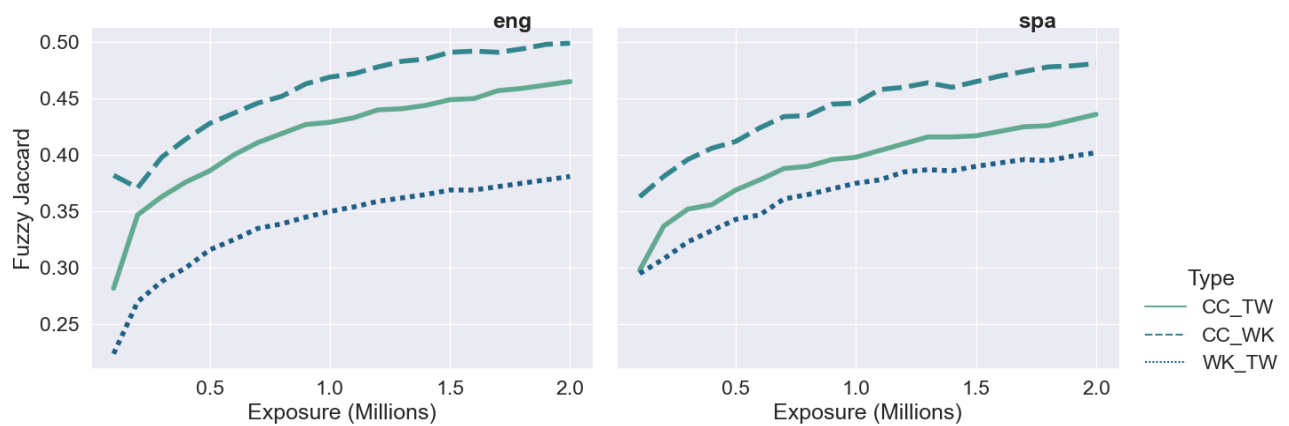

This figure shows, first, that there is a baseline similarity for each pair of registers. For example, social media and non-fiction articles are the least similar throughout. The ordering of registers is less important than the relative change within a register. In both languages, increased exposure leads to converging grammars. In other words, given a small amount of exposure, the two constructicons are quite different. But as the amount of exposure increases the constructicons settle onto a more similar set of constructions. These results are similar to those found in earlier experiments (Dunn and Tayyar Madabushi 2021). The figures for all 35 languages are available in the supplementary material.[5]

---

[5] https://doi.org/10.18710/CES0L8

The question is whether all languages converge across registers given increased exposure and, if they do converge, if they converge at the same rate. First, as an issue of quality control, three languages show no convergence for grammars drawn from the social media register (TW): Hindi, Tagalog, and Thai. The non-social media grammars for these languages do show convergence. For simplicity, here we focus on comparing the CC and WK registers. As shown in Figure 5, for most languages (with the three exceptions listed above) the trend in one register corresponds to the trend in the remaining registers.

Do languages from all families converge onto shared constructicons given increased exposure? This is shown in Figure 6, by family, with sub-plots similar to those in Figure 5. Each line represents the CC and WK grammars for a different language. First, the overall level of convergence is different by family. For example, Romance languages have more similar constructicons than Balto-Slavic languages. Second, and perhaps related, some families show much flatter convergence rates: for example, Uralic languages (like Hungarian and Finnish) do not show the same trend as other families. There is less convergence and the convergence which does take place is located a lower level of exposure. The opposite pattern is show in the Indo-Iranian family, where two languages show a very steep increase at lower levels of exposure (these are Hindi and Urdu).

The clear pattern here, however, is that register-specific grammars converge given increased exposure. While previous work focused on Germanic and Romance languages, we see this same pattern across a wider range of languages. There are exceptions, however, which show the importance an experimental framework that includes as many languages as possible. Register-specific constructicons have been exposed to unique sets of usage; we know that register is a significant cause of syntactic and lexical variation (Biber 2012). These constructicons are influenced by this variation, to different degrees. The fact that they converge given more exposure, however, indicates that the grammars are creating more general representations. Lexicons do not converge in this way, of course, and thus we see the reason for the difference in growth rates between grammars and lexicons: the lexicon does not acquire deeper generalizations given more exposure but the constructicon does.

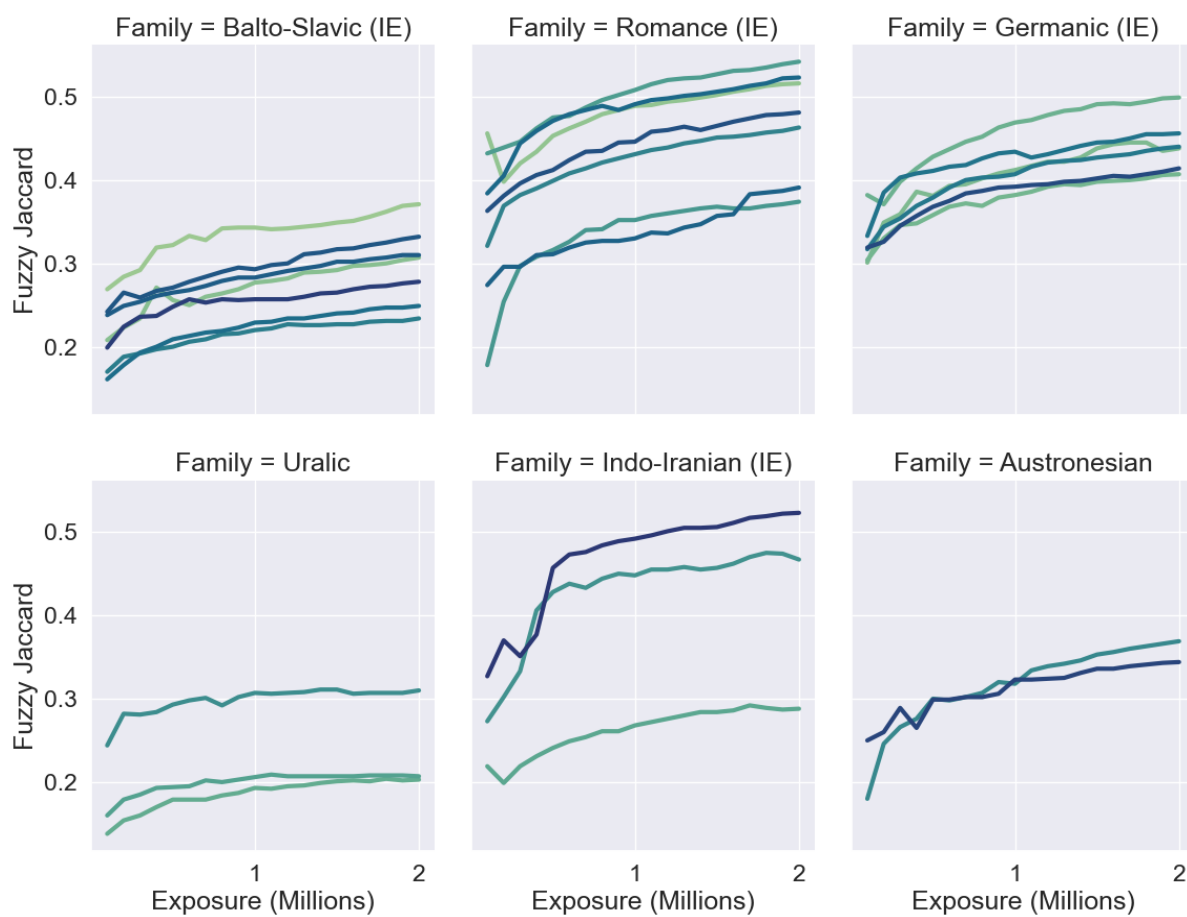

**Figure 6. Fuzzy Jaccard Similarity to Measure Convergence, by Family**

The second experiment asks whether there is a distinction between the entire constructicon (including infrequent constructions) and the core constructicon (only constructions that are frequent in an independent corpus). We measure convergence in the core constructicon using the weighted Jaccard similarity described above. Here we focus on the overall convergence rather than convergence given increased exposure.

The contrast is shown in Table 5 by family with the Weighted (left) and Fuzzy (right) similarity measures. The amount of exposure is fixed at 2 million words. As before, we focus on the web and non-fiction article registers for simplicity. In each case, the level of convergence is significantly higher when we focus on the core constructicon. Some language families have no internal consistency: for example, Balto-Slavic languages range from 0.44 (Czech) to 0.73 (Slovenian). There is no general trend within this family. For Romance languages, however, the range is much narrower: from 0.74 (Spanish) to 0.78 (Portuguese). No family shows higher internal agreement than Romance.

A few languages have particularly high convergence: for example, Urdu has a weighted similarity of 0.80, meaning that the core constructicon is quite similar across different registers. Thus, the impact of register variation is quite small. But other languages have quite low convergence: for example, Thai has a weighted similarity of 0.37, meaning that most of the core constructicon is not shared across registers. Thus, the impact of register variation is quite high.

**Table 5. Convergence in the Core (Weighted) and the Entire Constructicon (Fuzzy)**

| Name | Family | Weighted | Fuzzy |
|---|---|---|---|
| Indonesian | Austronesian | 0.73 | 0.37 |
| Tagalog | Austronesian | 0.57 | 0.34 |
| *Family Mean* | | *0.65* | *0.36* |
| Bulgarian | IE:Balto-Slavic | 0.65 | 0.37 |
| Czech | IE:Balto-Slavic | 0.44 | 0.31 |
| Latvian | IE:Balto-Slavic | 0.59 | 0.23 |
| Polish | IE:Balto-Slavic | 0.49 | 0.25 |
| Russian | IE:Balto-Slavic | 0.63 | 0.31 |
| Slovenian | IE:Balto-Slavic | 0.73 | 0.33 |
| Ukrainian | IE:Balto-Slavic | 0.50 | 0.28 |
| *Family Mean* | | *0.56* | *0.29* |
| Danish | IE:Germanic | 0.61 | 0.44 |
| Dutch | IE:Germanic | 0.72 | 0.44 |
| English | IE:Germanic | 0.69 | 0.50 |
| German | IE:Germanic | 0.73 | 0.41 |
| Norwegian | IE:Germanic | 0.71 | 0.46 |
| Swedish | IE:Germanic | 0.63 | 0.41 |
| *Family Mean* | | *0.68* | *0.44* |
| Farsi | IE:Indo-Iranian | 0.50 | 0.29 |
| Hindi | IE:Indo-Iranian | 0.64 | 0.47 |
| Urdu | IE:Indo-Iranian | 0.80 | 0.52 |
| *Family Mean* | | *0.65* | *0.43* |
| Catalan | IE:Romance | 0.75 | 0.52 |
| French | IE:Romance | 0.77 | 0.54 |
| Galician | IE:Romance | 0.77 | 0.37 |
| Italian | IE:Romance | 0.76 | 0.46 |
| Portuguese | IE:Romance | 0.78 | 0.52 |
| Romanian | IE:Romance | 0.76 | 0.39 |
| Spanish | IE:Romance | 0.74 | 0.48 |
| *Family Mean* | | *0.76* | *0.46* |
| Arabic | Semitic | 0.67 | 0.22 |
| Hebrew | Semitic | 0.59 | 0.12 |
| *Family Mean* | | *0.63* | *0.17* |
| Estonian | Uralic | 0.46 | 0.20 |
| Finnish | Uralic | 0.44 | 0.21 |
| Hungarian | Uralic | 0.55 | 0.31 |
| *Family Mean* | | *0.48* | *0.24* |
| Greek | Other | 0.69 | 0.43 |
| Korean | Other | 0.50 | 0.10 |
| Thai | Other | 0.37 | 0.38 |
| Turkish | Other | 0.61 | 0.15 |
| Vietnamese | Other | 0.54 | 0.37 |

Because the grammars are exposed to different registers, low agreement could be caused by higher grammatical variation across registers. It is also possible, in some cases, that this is caused instead by poor generalizations (for example, the absence of morphological parsing).

The question here has to do with the underlying difference between registers: if two contexts of production are quite different, it is expected that the grammars exposed to them should also be different. This is an experimental question for future work: is there a relationship between the linguistic distance between registers and the convergence of grammars exposed to those registers? The experiments here, however, do not control for underlying differences in register across these 35 languages (c.f., Li et al 2022).

The experiments in this section show that for most languages register-specific grammars converge when exposed to more usage. While the context of production is distinguished by unique lexical and syntactic patterns, the constructicon reaches more generalized constructions given more exposure and these constructions are more likely to be shared across registers. At the same time, the convergence of the core constructicon is much higher across all languages than the entire constructicon. Taken together, this means that the difference between registers is located in the long-tail of relatively infrequent constructions.

## 7 Unentrenchment Rates for Forgetting Constructions

This paper has experimented with the growth and convergence of the constructicon as computational CxG is exposed to more usage. This final set of experiments focuses on how constructions are forgotten or become unentrenched: how do we capture recency effects in exposure? After all, given the corpus-based approach taken in this paper, there is no limit to the amount of exposure that is possible: the corpora are quite large. Here we develop a simple model of unentrenchment that simulates the process by which constructions can be removed from the constructicon over time. In more practical terms, this part of the model prunes constructions from a grammar when they have not been recently observed.

Why would a model of the emergence of grammatical structure include the forgetting of that structure as well? First, from a computational perspective, recent work in NLP has developed increasingly precise distributional models of language which, at the same time, depend on enormous corpora (over 300 billion words: Bender et al. 2021). A computational model like the one presented in this paper, given modern corpora, could continue to learn from new exposure without practical limits. Thus, there is a need to explicitly limit the amount of time for which previous exposure remains directly available. Second, from a learning perspective, there is evidence that spacing between instances of exposure increases the degree to which an item is acquired (Vlach 2014). While this experimental paradigm operates under significantly different conditions than the model in this paper, the model as evaluated in Sections 5 and 6 is similar to testing a child for learning immediately after exposure has taken place. The model as evaluated in this section is similar to testing for learning after a delay, with an intermediate period of additional exposure (c.f., Vlach and De Brock 2019). In other words, does a delay after exposure (forgetting) lead to the increased entrenchment of constructions (c.f., Vlach 2019)?

We first merge the constructicons for each register at an exposure of 2 million words. This simulates a single learner who has been exposed to 2 million words from social media, from web pages, and from non-fiction articles. This merging creates a potentially large constructicon because the register-specific forms, themselves infrequent, are combined into a single grammar.

Second, we continue to observe unique sub-corpora from each register (CC, TW, WK) in increments of 100k words. Each construction in the grammar receives an activation weight with an initial value of 1. For each new sub-corpus in which a construction is not observed, its weight decays by 0.25. For each new sub-corpus in which a construction is observed in usage, its weight is returned to 1. When a construction's weight falls below 0, it is forgotten and removed from the grammar.

This simple model first accumulates constructions and then slowly forgets those constructions when they have not been recently observed. For simplicity, the unentrenching process crosses register boundaries, with each register being observed in turn. The degree of decay ensures that a construction specific to formal non-fiction articles, for example, would not be forgotten simply because it only occurs in non-fiction articles. Any construction that becomes unentrenched must be absent from multiple sub-corpora from each register.

How quickly are constructions forgotten? This is shown by family in Figure 7 which mimics the structure of Figure 6. The y-axis for each sub-plot is the size of the constructicon. At the beginning this is the merged grammar for each register. The x-axis is the amount of exposure, from 100k to 2 million words in increments of 100k. Here exposure is the same as in the previous experiments except that constructions are being forgotten rather than learned. As described above, each increment in which a construction is not observed decays that construction's weight until, finally, an unentrenched construction is removed from the grammar. This means that the learner has acquired the grammar given 2 million words of exposure and then forgets part of that grammar given another 2 million words of exposure.

**Figure 7. Rates of Unentrenchment by Language Family**

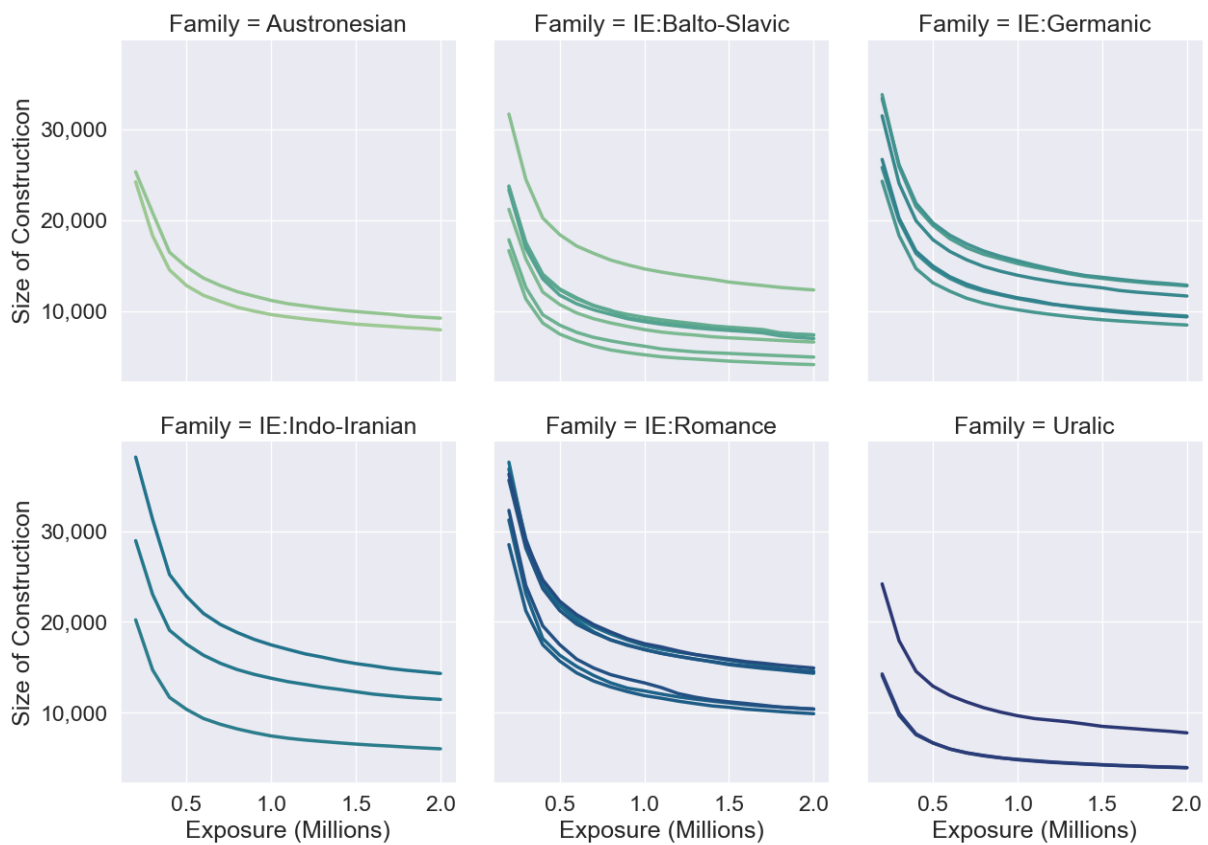

Figure 7 shows that the general pattern of unentrenchment is shared across languages: a sharp decline until approximately 500k or 600k words of exposure and then a slower rate as moderately frequent constructions are unentrenched. The most common core constructions remain in the grammar throughout. The Uralic family, which showed the lowest rate of convergence, also shows the lowest rate of unentrenchment. And Romance languages, with the highest rate of convergence, show the steepest reduction in the size of the constructicon. This implies that exposure is a mirrored process that influences both the emergence and the forgetting of grammars. Figures for each language are available in the supplementary material.[6]

Although there are patterns visible at the family level, there is not a significant relationship between the learning rate and forgetting rate at the level of individual languages. Figure 8 shows a regression between the growth rate of the constructicon (averaged across all three registers) and the unentrenchment rate, both measured using the α parameter as described above (i.e., both modelled using Heap's Law). The y-axis shows the learning rate, with languages toward the top with quickly growing constructicons. And the x-axis shows the unentrenchment rate, with languages to the right having quickly reducing constructicons. There is no clear relationship between the two at the level of individual languages, so that the growth rate cannot predict the rate of unentrenchment.

**Figure 8. Regression Analysis for the Learning and Forgetting Rates across Languages**

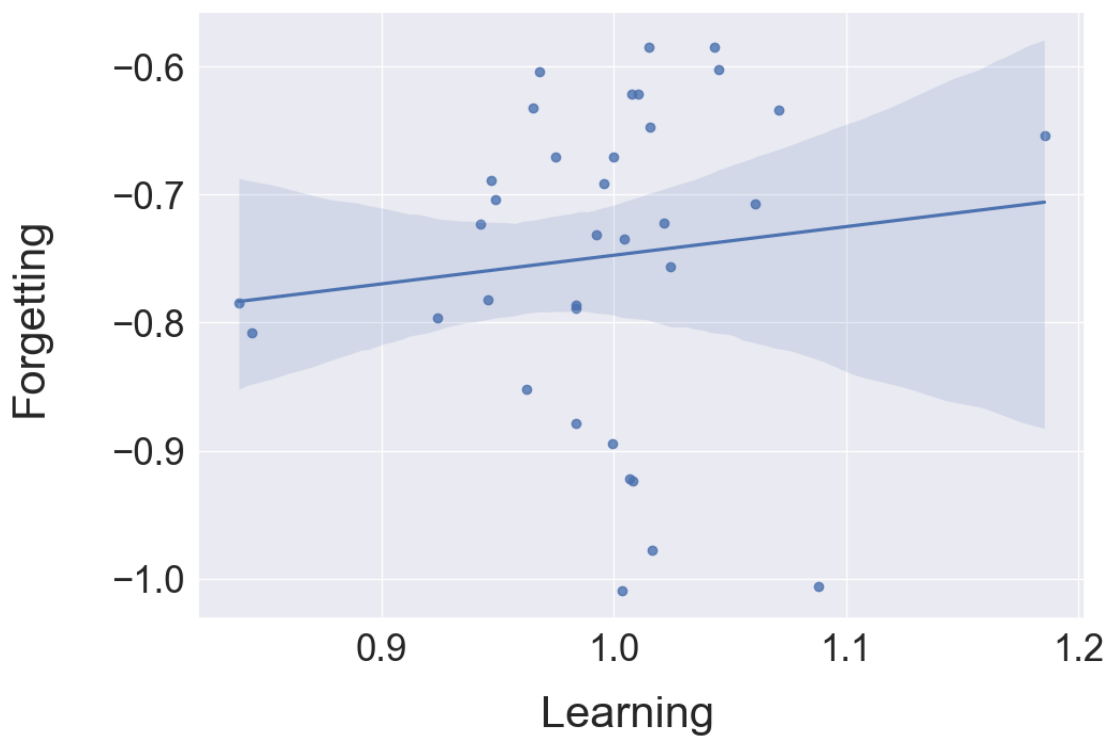

The exact rate of unentrenchment depends on parameters like the amount of weight decay and the increment size (here 100k) and the order in which registers are observed. Thus, the exact relationship between learning and forgetting in computational CxG is open to future research.

These experiments on unentrenchment use 2 million words of exposure across rotating registers to prune grammars, lowering the activation weight of a construction for each increment in which it does not appear. This pruning process is used to create a final constructicon that crosses register boundaries and retains only those constructions which have been continuously observed. These are shown in Table 6 by family, with the size of the final pruned constructicon for each language together with the merged lexicon (including the full 2 million words of exposure for each register). These grammars are released for all 35 languages.[7]

---

[6] https://doi.org/10.18710/CES0L8
[7] https://github.com/jonathandunn/c2xg/releases/tag/v1.0

*Table 6. Size of Lexicons Across Registers and Pruned Constructicons*

| Language | Code | Family | Lexicon | Constructicon |
|---|---|---|---|---|
| Indonesian | ind | Austronesian | 230,338 | 7,942 |
| Tagalog | tgl | Austronesian | 276,222 | 9,224 |
| | | *Family Mean* | *253,280* | *8,583* |
| Bulgarian | bul | IE:Balto-Slavic | 326,879 | 12,332 |
| Czech | ces | IE:Balto-Slavic | 433,558 | 6,595 |
| Latvian | lav | IE:Balto-Slavic | 416,373 | 4,110 |
| Polish | pol | IE:Balto-Slavic | 414,626 | 4,937 |
| Russian | rus | IE:Balto-Slavic | 426,785 | 7,394 |
| Slovenian | slv | IE:Balto-Slavic | 445,061 | 7,359 |
| Ukrainian | ukr | IE:Balto-Slavic | 445,751 | 6,984 |
| | | *Family Mean* | *415,576* | *7,102* |
| Danish | dan | IE:Germanic | 304,141 | 12,783 |
| Dutch | nld | IE:Germanic | 278,868 | 9,455 |
| English | eng | IE:Germanic | 186,285 | 12,856 |
| German | deu | IE:Germanic | 401,979 | 8,470 |
| Norwegian | nor | IE:Germanic | 392,655 | 11,663 |
| Swedish | swe | IE:Germanic | 339,043 | 9,368 |
| | | *Family Mean* | *317,162* | *10,766* |
| Farsi | fas | IE:Indo-Iranian | 263,120 | 6,028 |
| Hindi | hin | IE:Indo-Iranian | 233,432 | 11,468 |
| Urdu | urd | IE:Indo-Iranian | 171,992 | 14,325 |
| | | *Family Mean* | *222,848* | *10,607* |
| Catalan | cat | IE:Romance | 222,178 | 14,587 |
| French | fra | IE:Romance | 210,237 | 14,348 |
| Galician | glg | IE:Romance | 269,244 | 10,419 |
| Italian | ita | IE:Romance | 251,608 | 9,897 |
| Portuguese | por | IE:Romance | 206,982 | 14,461 |
| Romanian | ron | IE:Romance | 309,015 | 10,396 |
| Spanish | spa | IE:Romance | 210,276 | 14,929 |
| | | *Family Mean* | *239,934* | *12,720* |
| Arabic | ara | Semitic | 473,857 | 4,181 |
| Hebrew | heb | Semitic | 366,149 | 3,322 |
| | | *Family Mean* | *420,003* | *3,752* |
| Estonian | est | Uralic | 732,983 | 3,933 |
| Finnish | fin | Uralic | 746,945 | 3,962 |
| Hungarian | hun | Uralic | 611,732 | 7,787 |
| | | *Family Mean* | *697,220* | *5,227* |
| Greek | ell | Other | 337,543 | 11,212 |
| Korean | kor | Other | 1,148,698 | 2,302 |
| Thai | tha | Other | 98,794 | 4,875 |
| Turkish | tur | Other | 525,719 | 3,184 |
| Vietnamese | vie | Other | 101,190 | 5,649 |

# 7 Discussion and Conclusions

While usage-based grammar hypothesizes a relationship between exposure and the emergence of constructions, a large gap has remained in the literature: no previous research actually predicted grammars given exposure. This gap has meant that the usage-based hypothesis has relied on small-scale studies of isolated constructions to model exposure and emergence. This paper has augmented these small-scale studies with computational models of the emergence of grammar across 35 languages, with three distinct contexts of production providing unique sets of exposure. The goal has been to experiment with the relationship between exposure and emergence: (i) comparing the growth of the lexicon with the growth of the grammar, (ii) measuring the convergence of grammars exposed to unique registers, and (iii) measuring how constructions can become unentrenched given more exposure in which they are not observed.

The first set of experiments has shown that the lexicon grows significantly more quickly than the constructicon. Closer examination shows that the lexicon and the constructicon for each language grow at the same or similar rates across all three registers; this shows that the difference is not an arbitrary property of one register or language. Further, there is no significant relationship in most cases between the growth curve of the lexicon and the growth of the constructicon. In other words, the constructicon is not simply an extension of the lexicon. In part this is because the nature of constructions can change given more exposure: they become more abstract and less idiomatic.

The second set of experiments has shown that constructicons converge across registers given increased exposure. In other words, grammars exposed to 2 million words are more similar than those exposed to 1 million words, which are in turn more similar than those exposed to 500k words. This is true across languages and across registers, although the convergence rate varies by language family. Further, the core constructicon, as defined by frequency of constructions, has higher agreement across registers than the entire constructicon. This means that there is a shared core of constructions across registers. Grammars become more similar given increased exposure because their constructions become more generalized. This shows why the constructicon grows more slowly than the lexicon: given more exposure, more abstract constructions are acquired and these are shared across contexts of production.

The third set of experiments reverses the learning process by allowing constructions to be forgotten over another round of exposure (again 2 million words in increments of 100k). Each construction is given an activation weight which slowly decays when it has not been recently observed. Languages show a similar rate of unentrenchment; while this rate does not have a significant relationship with the growth rate within each language, it does show that exposure can reasonably be seen as part of both the learning and the forgetting of constructions.

The wider contribution of this paper is to formulate a computational model of usage-based grammar that predicts constructicons given exposure to actual usage. These constructicons are then available for computational corpus analysis. For example, these types of grammar have been shown to be effective at capturing differences between regional dialects. This sort of external evaluation provides additional evidence that this model of usage-based construction grammar provides both theoretical insights and practical tools. A Python package for using these constructicons is available on GitHub.[8]

How do these computational experiments relate to the emergence of a shared constructicon within an actual population of speaker-hearers? In other words, this model focuses on exposure as perception but ignores the interplay between perception and production (c.f., Dunn and Nini 2021). From a different perspective, the exposure in this paper is contributed by corpora which were not necessarily produced from a single limited speech community. We know from other recent computational work, however, that the constructicon can also be used to model spatial variation in language with a high degree of accuracy (Dunn 2018b, 2019b; Dunn and Wong 2022). In work like this, spatial variation is an approximation for capturing different populations of speaker-hearers. In other words, a computational model of the dialect of New Zealand English in such work is a combination of (i) a usage-based constructicon and (ii) weights for each construction which indicate the preferences of the population of speaker-hearers in New Zealand.

---

[8] https://github.com/jonathandunn/c2xg/releases/tag/v1.0

The next question, for future work, is whether a model of the emergence of grammatical structure that is situated within specific populations (as in Dunn 2018b, 2019b) would lead to observed dialectal differences as an implicit side-effect of emergence rather than as an explicit discriminative model. In other words, this previous work on computational dialectology has assumed an almost omniscient observation of large corpora representing each dialect and then used discriminative models (i.e., classifiers) to distinguish between dialects. The emergence of dialects, however, cannot take place in this manner because a community of speaker-hearers does not have this same omniscient information. If a model of emergence were instead situated within each specific population, a sort of usage-based dialectology, would the same dialect-specific grammars result?

Finally, what can computational experiments tell us about the specific role of exposure in the emergence of syntactic structures? For example, we know that one of the factors behind lexical development within individuals comes from different amounts of exposure (Huttenlocher et al. 1991). As with the experiments in Section 4, that work focused on the amount of exposure and the number of vocabulary items learned. The replication of such studies with constructions rather than simple lexical items is made challenging without computational methods because it requires positing a constructicon in a reproducible manner. With access to computational methods, however, we know that the production of individuals within a highly constrained register remains unique (Dunn and Nini 2021), a fact likely related to the unique exposure these individuals have experienced. The model in this paper has made the simplifying assumption that the grammar at each increment of exposure is independent; thus, we can measure the hypothesized growth of the constructicon, but not the scaffolding or boot-strapping of structure during acquisition (c.f., Conboy and Thal 2006).

For example, in this kind of model the output for each iteration of exposure would serve as the slot-constraints for emerging structure in the next iteration: until we have acquired a particular class of verbs, for example, no constructions which depend on that class could emerge. Thus, the results in this paper account for the amount of exposure but not the *order* of exposure. For CxG, the order depends not just on the specific corpora being observed but on the existing grammar itself as a starting point for more complex representations. The next question, again for future work, is whether a model with scaffolded structures like this can explain the convergence rates observed in Section 6, which clearly distinguish between the core and the periphery of the grammar.

# Appendix 1: Glossary of Computational Terms

*Potential Construction*. Given an observed string, the learner could hypothesize many competing structural analyses. From the perspective of CxG, these structures could involve boundaries, slot types, and slot constraints. Part of modelling the emergence of such structures is to hypothesize what possible constructions are being observed in a particular set of input.

*Hypothesis Space*. A constructicon in this model is a set of constructions, with each construction a sequence of slot-constraints. Given a specific language, the set of possible constructions depends both on the formation of categories (e.g., what semantic or syntactic categories are recognized) as well as the usage observed in the corpora. Richer representations like CxG have a larger hypothesis space because they contain more potential structures.

*ΔP*. Within the CxG model, the distribution of sequences is quantified using association measures, specifically this ΔP which captures the probability of an outcome given a cue. In this case, the cue is a particular linguistic sequence and the outcome is the following sequence. For example, sequences with many possible following items will have lower values.

*Skip-Gram Negative-Sampling*. The semantic domains for each language are formed using word embeddings trained using the SGNS method (within the fastText framework). This method essentially trains a logistic regression classifier to predict the most likely context words for each target word in a corpus. Work in NLP has shown that such embeddings are closely related to matrices of association measures.

*Slot-Constraint*. Constructions are based on symbolic constraints, in which an utterance is produced by some series of specific slot-constraints. A *slot* in this sense is a segmentation (here, word-level segmentation) and a *constraint* is the type of category used to formulate the generalization.

*Slot-Fillers*. In this model, there are three types of slot-fillers: lexical, syntactic, and semantic. While CxG could formulate joint semantic-syntactic constraints, the constructions learned in this paper show that implicit relationships between constraints emerge during learning.

*Minimum Description Length* (MDL). Given a grammar, we evaluate its quality by describing how well it fits or describes a particular test corpus. For language models, the measure most often used is perplexity, for which a better model will provide a smaller description of the data. MDL is a further refinement of perplexity which takes model complexity into account as well. The resulting trade-off is a balance between memory (i.e., storing all possible constructions) and computation (i.e., relying on fully compositional phrase structure rules).

*Starting Node*. Within the beam-search algorithm for finding potential constructions, each hypothesized structure must have boundaries. The starting node refers to the beginning of such a structure, the point from which the search begins.

*Candidate Stack*. From each starting node, a very large number of potential representations (potential constructions) is possible. Rather than maintaining all such candidates, we store in short-term memory the best options and then prune or evaluate these candidates using a beam-search.

**Data Availability Statement**

The full code to reproduce these experiments is made available in a Python package.[9] The corpora used for learning grammars are contained in the supplementary, when possible given the terms of use. In addition, the supplementary material[10] contains the following items:

(1) Semantic domains for each language

(2) Constructicon with examples for each language

(3) Growth curve figures for 35 languages

(4) Regression results for growth curve estimation

(5) Grammar convergence figures for 35 languages

(6) Unentrenchment rate figures for 35 languages

(7) Complete raw results and scripts for reproducing all models and figures

---

[9] https://github.com/jonathandunn/c2xg/releases/tag/v1.0
[10] https://doi.org/10.18710/CES0L8